%% file: main.tex
\documentclass[10pt,twocolumn,letterpaper]{article}

\usepackage[pagenumbers]{cvpr} 


\definecolor{cvprblue}{rgb}{0.21,0.49,0.74}
\usepackage[pagebackref,breaklinks,colorlinks,allcolors=cvprblue]{hyperref}
\usepackage{multirow}
\usepackage[accsupp]{axessibility}

\def\paperID{693} 
\def\confName{CVPR}
\def\confYear{2025}

\title{Empowering Large Language Models with 3D Situation Awareness}

\author{Zhihao Yuan$^{1,2}$, Yibo Peng$^{1,2}$, Jinke Ren$^{1,2}$, Yinghong Liao$^{1,2}$, Yatong Han$^{1}$, Chun-Mei Feng$^{3}$, \\ Hengshuang Zhao$^{4}$, Guanbin Li$^{5}$, Shuguang Cui$^{2,1}$, Zhen Li$^{2,1}$\thanks{Corresponding author.} \\
\\
$^{1}$ FNii-Shenzhen, CUHKSZ~~~
$^{2}$ SSE, CUHKSZ~~~
$^{3}$ IHPC, A*STAR, Singapore~~~
$^{4}$ HKU~~~
$^{5}$ SYSU 
}

\begin{document}
\maketitle
\input{sec/0_abstract}

\input{sec/1_intro}

\input{sec/2_related}
\input{sec/3_method}

\input{sec/4_experiment}

\input{sec/5_conclusion}
\input{sec/6_ac}

{
    \small
    \bibliographystyle{ieeenat_fullname}
    \bibliography{main}
}

\input{sec/X_suppl}


\end{document}

%% file: sec/0_abstract.tex
\begin{abstract}

Driven by the great success of Large Language Models (LLMs) in the 2D image domain, their application in 3D scene understanding has emerged as a new trend. A key difference between 3D and 2D is that the situation of an egocentric observer in 3D scenes can change, resulting in different descriptions (e.g., ``left" or ``right"). However, current LLM-based methods overlook the egocentric perspective and  use datasets from a global viewpoint. To address this issue, we propose a novel approach to automatically generate a situation-aware dataset by leveraging the scanning trajectory during data collection and utilizing Vision-Language Models (VLMs) to produce high-quality captions and question-answer pairs. Furthermore, we introduce a situation grounding module to explicitly predict the position and orientation of the observer's viewpoint, thereby enabling LLMs to ground situation descriptions in 3D scenes. We evaluate our approach on several benchmarks, demonstrating that our method effectively enhances the 3D situational awareness of LLMs while significantly expanding existing datasets and reducing manual effort.

\end{abstract}

%% file: sec/1_intro.tex
\section{Introduction}
\label{sec:intro}

\begin{figure}[t]
  \centering
   \includegraphics[width=1.0\linewidth]{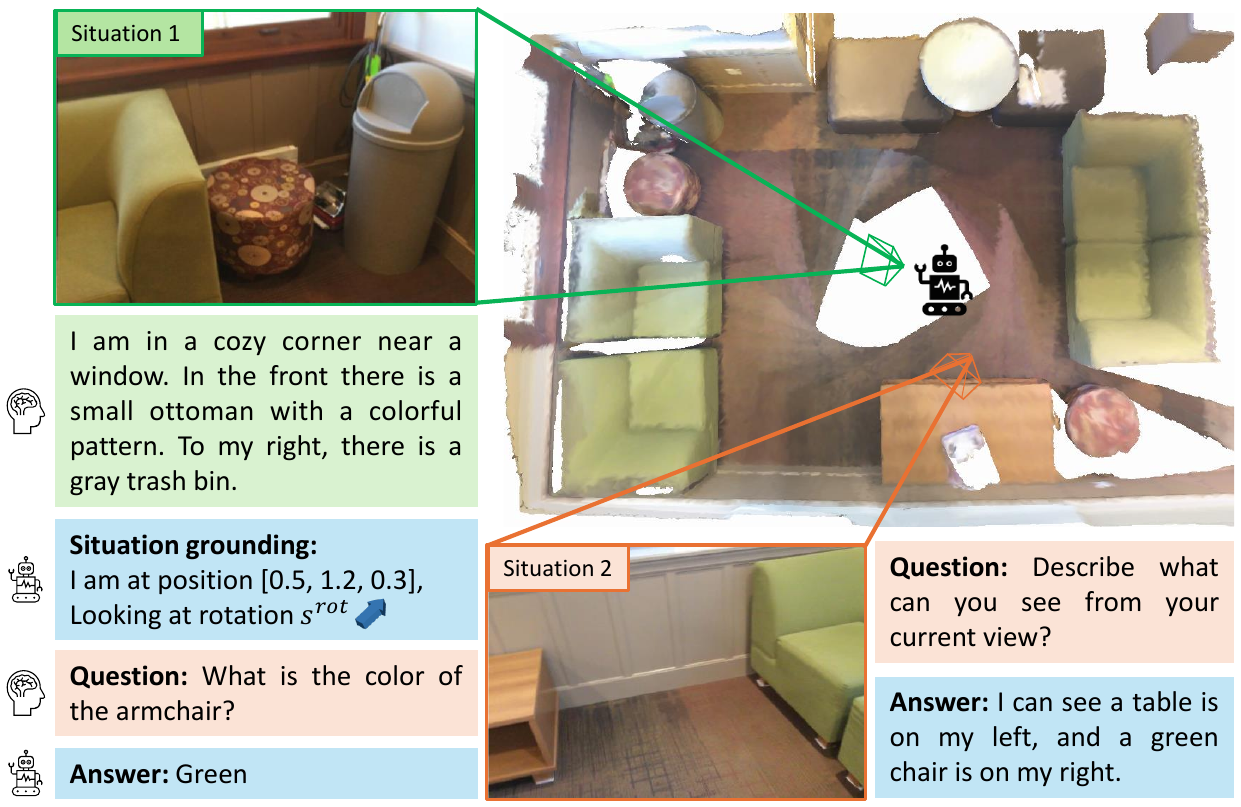}

   \caption{Illustration of 3D Situation Awareness. The LLMs can accurately ground situation descriptions to the observer's  position and orientation, enabling context-aware question answering based on the observer's viewpoint.}
   \label{fig:1}
\end{figure}

Recently, Large Language Models (LLMs)~\cite{touvron2023llama, team2024gemma, OpenAI_2023} have revolutionized natural language processing, showcasing remarkable capabilities in image understanding tasks~\cite{liu2024visual, krishna2017visual, gpt4o}, such as image captioning, visual question answering (VQA), and engaging in dialogs about visual content. Building on this success, researchers have begun to explore the use of LLMs in the three-dimensional (3D) domain~\cite{hong20233d, xu2024pointllm, chen2024ll3da}, aiming to bridge the gap between language and 3D visual data. In 3D vision and language tasks—including 3D visual grounding, 3D captioning, and 3D VQA—the objective is to unify these tasks under the framework of next-token prediction. To achieve this, researchers focus on learning a 3D representation that aligns with pre-trained text embedding spaces, enabling seamless integration with language models.

However, a significant challenge in applying LLMs to 3D tasks is the scarcity of annotated 3D-text data, which is crucial for training such models effectively. Unlike 2D images, which have abundant paired data in the form of captions and annotations, 3D data lacks extensive textual descriptions. To overcome this limitation, researchers have explored methods to generate additional  3D-text data to augment the training process. For example, 3D-LLM~\cite{hong20233d} utilizes multi-view images of 3D scans to generate captions, leveraging rich visual information from different perspectives. Similarly, LEO~\cite{huang2023embodied} and SceneVerse~\cite{jia2024sceneverse} employ scene graphs and harness the capabilities of ChatGPT to generate captions and question-answer (QA) pairs, enriching the textual annotations associated with 3D content.

Despite these efforts, inherent limitations exist in the current data generation processes. Firstly, most approaches to 3D scene understanding emphasize global perspectives, providing comprehensive overviews of environments while overlooking the importance of situational contexts—specific viewpoints or scenarios within a scene crucial for accurate interpretation. Unlike static images with fixed viewpoints, 3D scenes are dynamic, and the perceived situation can change when an embodied agent moves through the environment. As illustrated in Fig.~\ref{fig:1}, an object like a sofa may appear on the left side from one viewpoint (Situation 1) but on the right side from another viewpoint (Situation 2). Without specifying the situational context, such variations can lead to ambiguity during model training and degrade performance in tasks that require spatial understanding.

Moreover, current scene graph-based methods \cite{hong20233d, huang2023embodied, jia2024sceneverse, clevr3d} for data generation rely heavily on ground-truth 3D instance labels to construct accurate scene graphs. Acquiring these 3D labels is labor-intensive and costly and limits the scalability of the data generation process. Existing labeled datasets are often insufficient, lacking coverage of all objects within a scene, especially small objects and those belonging to rare categories. Additionally, the relationships between objects are typically predefined using fixed templates, which restricts the ability of models to handle open vocabulary scenarios and capture the richness of the real-world.

To overcome these challenges, we propose a novel approach to automatically generate a situational dataset, termed \textbf{View2Cap}. Our key insight is that 3D scans are commonly reconstructed from RGB-D videos, where the camera trajectory inherently represents an egocentric exploration of the environment by a human observer. By leveraging this naturalistic data, we can capture the situational context not included in existing datasets. Specifically, we utilize 2D Vision-Language Models (VLMs) to generate captions and QA pairs from individual frames of the RGB-D videos. This approach effectively distills knowledge from well-established 2D models into the 3D domain, capitalizing on the strengths of 2D VLMs. Concurrently, we record the camera pose of each frame, which, in combination with the depth information, allows us to extract the corresponding point cloud for that particular region. This methodology enables us to create a point cloud-text dataset with situational context, capturing the dynamic perspectives encountered by an embodied agent moving through a 3D environment. Importantly, this strategy reduces data generation costs and supports the scalable creation of datasets capable of handling open vocabulary scenarios without the need for extensive manual annotations or 3D labels.

In addition to the dataset creation, enhancing the situational understanding of LLMs requires models that can explicitly ground descriptions in the 3D space. To this end, we propose a \textbf{Situation Grounding (SG)} module that builds upon existing 3D LLM architectures. This module allows the model to predict situational positions and view rotations based on textual descriptions and the scene-level point cloud. By treating each object within the scene as an anchor point, the model can predict the distance and angle class offsets relative to the observer's viewpoint. This formulation transforms the complex problem of pose estimation into a more tractable classification task, simplifying the learning process and improving the model's ability to comprehend and reason about spatial relationships in 3D environments.


Our contributions can be summarized as follows:

\begin{itemize}
    \item We introduce \textbf{View2Cap}, a scalable 3D dataset that provides paired data of situational positions, rotations, region point clouds, textual descriptions, and QA pairs. This dataset is generated automatically without the need for 3D labels or extensive manual annotations, enabling the study of situational context in 3D scene understanding.
    \item We propose the Situation Grounding \textbf{SG} module, that can be integrated into existing LLMs, allowing for explicit grounding of situational descriptions in 3D scenes. This module transforms pose estimation into a classification problem, facilitating easier training and improved spatial reasoning.
    \item Through extensive experiments, we demonstrate that combining our situational dataset and grounding module significantly enhances the situational awareness and performance of existing 3D LLMs in various tasks.
\end{itemize}

%% file: sec/2_related.tex
\section{Related Work}
\label{sec:related}

\noindent\textbf{Indoor 3D Scene Understanding.} 3D scene understanding involves perceiving and interacting with 3D environments, encompassing tasks such as 3D segmentation \cite{schult2023mask3d}, 3D visual grounding \cite{chen2020scanrefer, yuan2021instancerefer, yuan2022toward}, 3D captioning \cite{chen2021scan2cap, yuan2022x}, and 3D visual question answering (VQA) \cite{clevr3d, ma2022sqa3d}. The development of large-scale RGB-D scan datasets has significantly advanced this field. Notably, ScanNet~\cite{scannet} provides extensive annotations for indoor scenes, facilitating research in 3D scene understanding. Matterport3D~\cite{chang2017matterport3d} offers a large-scale collection of richly annotated house-level environments, providing high-resolution RGB-D scans and detailed semantic labels, which have been instrumental in advancing tasks such as 3D reconstruction, navigation, and semantic understanding. EmbodiedScan~\cite{wang2024embodiedscan} enriches these annotations by providing more fine-grained object bounding boxes, particularly focusing on small objects and diverse class labels, with assistance from models like SAM~\cite{kirillov2023segment}.

\noindent\textbf{Large Language Models in 3D.} Inspired by the success of LLMs in image understanding, researchers have begun exploring the integration of 3D inputs with LLMs to leverage their impressive reasoning and generalization capabilities for 3D understanding~\cite{yuan2024visual}. Models such as PointLLM~\cite{xu2024pointllm} and GPT4Point~\cite{qi2024gpt4point} attempt to map point clouds into the token space of LLMs to generate captions for objects. However, they struggle to handle scene-level point clouds due to the complexity of indoor environments. 3D-LLM~\cite{hong20233d} pioneers the use of LLMs for scene understanding but still relies on 2D features.
Recent models like LL3DA~\cite{chen2024ll3da}, Chat-3D~\cite{huang2023chat}, and LEO~\cite{huang2023embodied} have investigated the use of 3D encoders for scene-level tasks. To enhance grounding abilities, Grounded 3D-LLM~\cite{chen2024grounded} introduces referent tokens and employs contrastive learning to unify grounding with textual responses. Similarly, Chat-3D~\cite{huang2023chat} proposes the use of object identifiers (object IDs) to facilitate referring expressions and grounding mechanisms. However, these models still lack a comprehensive understanding of situation awareness in 3D space.

\noindent\textbf{Situation Awareness in 3D Space.} A key difference between 2D and 3D scene understanding lies in situation awareness. In 2D images, the viewpoint is fixed, making spatial relationships like left and right straightforward to determine. In contrast, in 3D spaces, these relationships can change with the observer's position. For example, left/right relationships can reverse when moving from one side of a room to another. Some works have addressed this problem using data augmentation techniques, such as MVT~\cite{huang2022multi} and ViewRefer~\cite{guo2023viewrefer}. SQA3D~\cite{ma2022sqa3d} first proposes to describe the situation in text and then conduct tasks like VQA. However, it relies on human annotators to write these descriptions, making it costly and challenging to scale for the large-scale training required by 3D LLMs.
Our method addresses this limitation by using an automatic situation dataset generation pipeline that leverages the capabilities of 2D VLMs and the trajectories inherent in RGB-D dataset collection processes. 

%% file: sec/3_method.tex
\section{Method}
\label{sec:method}

To enable LLMs to comprehend situational contexts within 3D spaces, we propose a method that involves an automatic data generation pipeline for building a situation dataset and a novel module for situation grounding.

\subsection{Situation Dataset}

\noindent\textbf{Automatic Data Generation.}
%
Our data generation process leverages the natural exploratory behavior inherent in RGB-D videos, which serves as first-person navigations through 3D environments. For each video frame, we extract the situational context directly from the camera extrinsic, capturing precise positional and rotational information. Moreover, we employ VLMs to generate captions from the corresponding 2D images extracted from the video frames. It provides elaborate information about the environment beyond 3D labels. Additionally, we utilize VLMs to generate QA pairs related to each situation, providing more specific and direct supervision compared to captions alone. Specifically, we use Llava-onevision~\cite{li2024llava} as VLM for its comprehensive caption ability and open-sourced.

For situation descriptions, we generate two types of captions: simple and detailed. The simple captions focus on the primary objects and their relationships within the scene, ensuring that the model grasps the essential components of the environment. The detailed captions provide an elaborate account of all visual information presented in the image, including background elements like the floor and overall environment, as shown in Fig.~\ref{fig:4}.

For the situation QA pairs, we define four categories of questions: (1) object identification, prompting recognition of objects presented in the image; (2) spatial relationships, describing the positions of objects relative to each other from the viewer's perspective (e.g., left, right, front, behind); (3) visual features, showing distinctive attributes such as colors, shapes, sizes, and textures; (4) insights into the overall layout of the room, enhancing the model's understanding of the scene as a whole.

\begin{table}[t]
\centering
\small
\begin{tabular}{lcc}
\toprule
Statistic                                & SQA3D     & View2Cap \\ \midrule
Total $s^{\text{txt}}$                   & 20,369    & 231,184 \\ 
Total $q$                                & 33,403    & 553,779 \\ 
Unique $q$                               & 26,091    & 92,877 \\ \midrule
Total scenes                             & 650       &  2,841  \\ 
Total objects                            & 14,925    &  71,376 \\ \midrule
Average $s^{\text{txt}}$ length          & 17.49     &  54.73 \\ 
Average $q$ length                       & 10.49     &   8.55 \\ 
Average $a$ length                       & 1.10 & 5.74 \\ \bottomrule
\end{tabular}
\caption{Dataset Statistics for SQA3D and View2Cap.}
\end{table}

\begin{figure*}[t]
  \centering
   \includegraphics[width=1.0\linewidth]{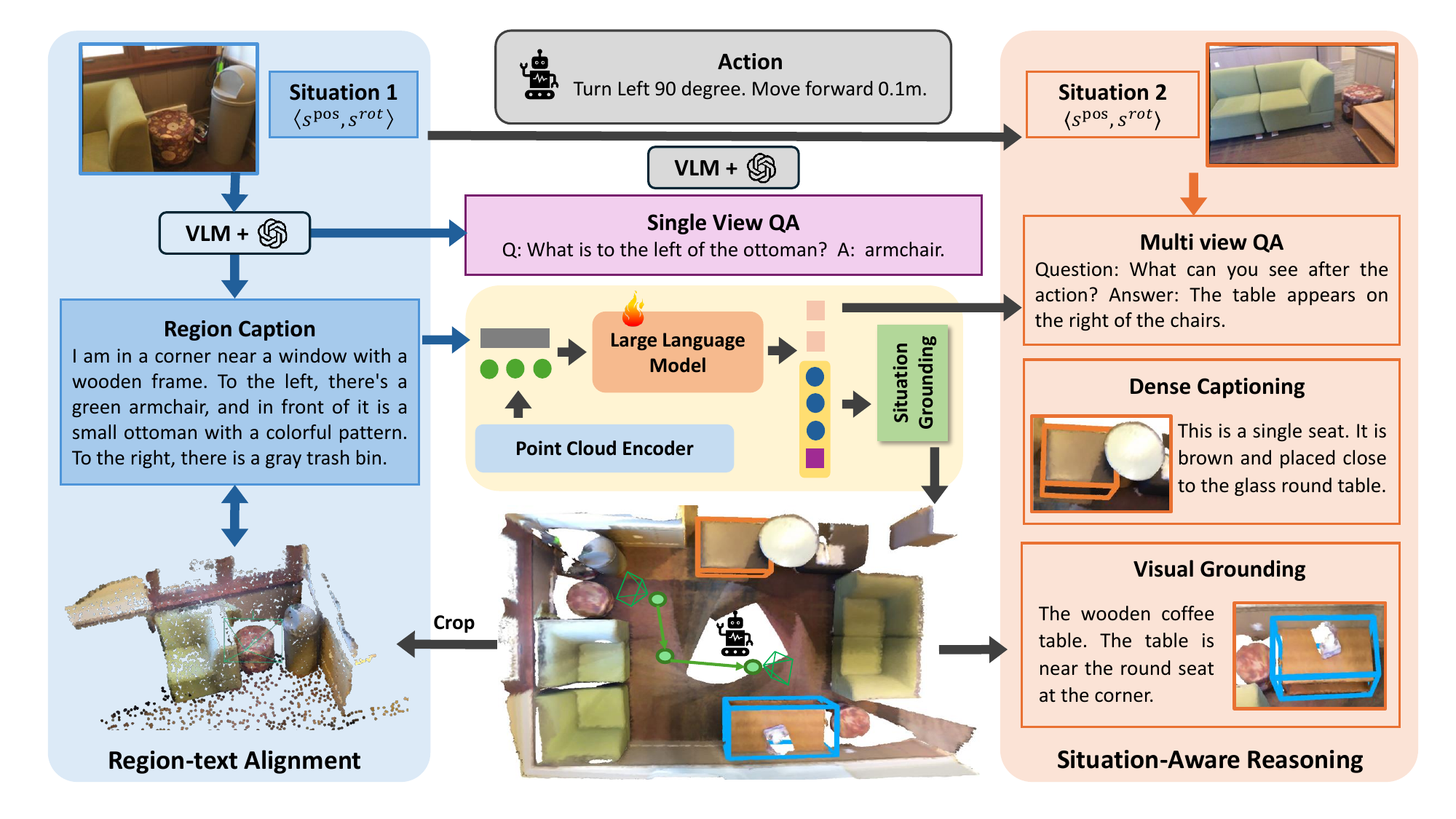}

   \caption{\textbf{Overview of our method.} The left part illustrates the process of region-text alignment. Paired point cloud and caption data are generated using VLM and RGB-D videos. The LLM is fine-tuned to align features from the point cloud encoder and generated region caption. For situation grounding, the region caption is fed into the LLM to predict the viewpoint of the observer \(\mathbf{s}^{pos}\) and \(\mathbf{s}^{rot}\). The right part shows the situation-aware instruction tuning process, where QA data is generated using multi-view images and corresponding actions.}
   \label{fig:2}
\end{figure*}


\noindent\textbf{Dataset Verification and Refinement.}
To ensure the quality and reliability of the generated dataset, we implement a verification and refinement process leveraging GPT-4. 
Specifically, we utilize the 3D labels of point clouds and employ GPT-4 to evaluate whether the generated captions include all objects present in the scene. We use annotations from EmbodiedScan \cite{wang2024embodiedscan}, as it annotate more fine-grained categories and small objects compared with the labels from the original scan datasets. The scoring criteria are adopted from PointLLM~\cite{xu2024pointllm}, including correctness, hallucination, and general considerations, range from 0-5. The average score for View2Cap is 3.09, indicating that captions derived from 2D images can effectively serve as high-quality supervision for training 3D models. We also ask GPT-4 to refine the validation set based on the labels, which increases the average score to 3.31. This refinement step enhances the reliability of our evaluations when assessing 3D models on validation sets. Detailed information about our scoring process is provided in the Supplementary Material.

For the ViewQA dataset, we aim to enrich the diversity and informativeness of the QA pairs. For each image, the VLM generates 10 questions for each predefined type. We then utilize GPT4 to rank these QA pairs based on criteria such as relevance, clarity, and informativeness. Items that fall below a certain threshold are excluded from the dataset. This ranking and filtering process ensures that only the most pertinent and well-constructed QA pairs are retained, thereby enhancing the overall quality of the dataset.

In total, we collect more than 200K situation descriptions and 550K QA pairs using 2,841 scans from ScanNet \cite{chen2021scan2cap}, 3RScan \cite{wald2019rio}, and Matterport3D \cite {chang2017matterport3d} datasets, which is 10x large than the SQA \cite{ma2022sqa3d}. Our average situation text length (54.73) is much longer than SQA3D (17.49). Our average answer length of 5.74 is also longer. More detailed statistics are provided in the Supplementary Material.

\subsection{Model Architecture}

Our model takes as input a point cloud \( \mathbf{P} \in \mathbb{R}^{N \times 6} \), representing a 3D scan where each point includes spatial coordinates and RGB color information. Additionally, it processes a situation description \( s \) and a task instruction \( t \). The goal is to generate the task answer \( a \). In addition, we ask the model to predict the situation position \( \mathbf{s}^{\text{pos}} = (x, y, z) \in \mathbb{R}^3 \) and the rotation of the front view direction represented as a quaternion \( \mathbf{s}^{\text{rot}} = (q_x, q_y, q_z, w) \), where \( q_x, q_y, \) and \( q_z \) are the vector components representing the axis of rotation scaled by  \( w \).

\noindent\textbf{Point Cloud Encoder.} The point cloud \( \mathbf{P} \) is segmented into \( K \) instances \( \{ \mathbf{P}_k \} \), each representing an object in the scene. For each instance, we group the points into local patches \( \{ \mathbf{p}_{k,l} \}_{l=1}^{L_k} \), where \( L_k \) is the number of patches in instance \( k \). Each patch contains neighboring points grouped based on spatial proximity.
The patch sequence along with a special token [CLS] is processed by a Vision Transformer (ViT) \cite{dosovitskiy2020image} to obtain instance feature $\mathbf{v}_k$. The point cloud encoder~\cite{zhou2023uni3d} is fine-tuned on     a large-scale 3D object dataset~\cite{deitke2023objaverse} in the classification task.

\noindent\textbf{Connector.} To enhance the spatial relationships between objects \cite{huang2023embodied, huang2023chat}, we employ spatial attention layers~\cite{chen2022language} to fuse the coordinates of objects with their semantic features. This fusion effectively integrates spatial and semantic information, which is crucial for comprehensive scene understanding. Subsequently, to map the instance-level features into the embedding space of the LLM, we utilize a simple multilayer perceptron (MLP) $f_{\text{proj}}$ to obtain the processed visual tokens corresponding to the object instances in the scene:
\begin{equation}
    \tilde{\mathbf{v}}_k = f_{\text{proj}} (\mathbf{v}_k).
\end{equation}
This projection ensures that the visual features are compatible with the LLM's embedding space.



\noindent\textbf{Large Language Model.}
Following multimodal LLM approaches, we tokenize the visual features and interleave them with text tokens to serve as input to the LLM. Specifically, the input sequence begins with system messages and the situation description \( t_s \), followed by the visual tokens, and concludes with the task instruction \( t \):
\begin{equation}
\text{Input} = [ t_s, \tilde{\mathbf{v}}_1, \tilde{\mathbf{v}}_2, \dots, \tilde{\mathbf{v}}_K, t ].
\end{equation}
This structure allows the LLM to process the visual context within the flow of textual information.
During the forward pass, the LLM processes the entire input sequence and generates hidden states for each token. We introduce a special grounding token \([ \text{GRD} ]\) in the output sequence. Let \( \{\mathbf{h}_k\} \) represent the hidden state corresponding to the visual tokens \( \{ \tilde{\mathbf{v}}_k \} \) and \( \mathbf{h}_{\text{GRD}} \) as the hidden state corresponding to this grounding token.

\begin{figure}[t]
  \centering
   \includegraphics[width=0.8\linewidth]{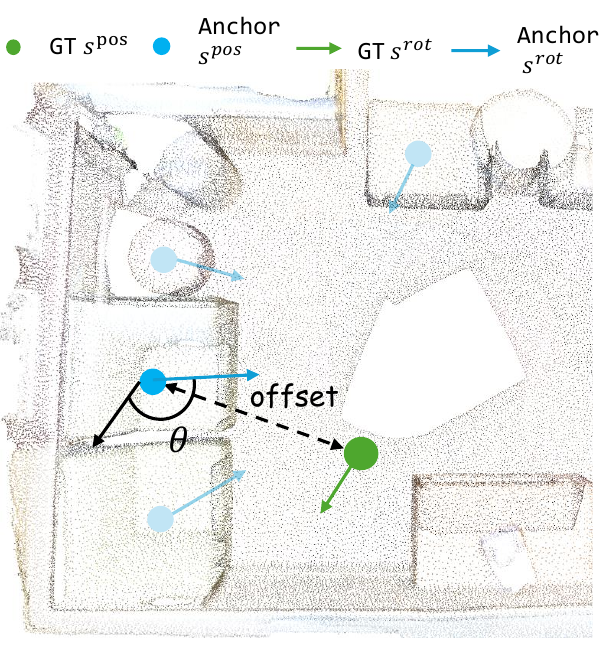}

   \caption{\textbf{Situation prediciton.} We treat each object as anchor (\textcolor[RGB]{15, 158, 213}{blue}), where the center is \( \mathbf{a}_k^{\text{pos}} \).  We set each \( \mathbf{a}_k^{\text{rot}} \) point to the center of the room. Ground truth position and rotation are shown in \textcolor[RGB]{78, 167, 46}{green}. The dotted line shows the offset from \( \mathbf{a}_k^{\text{pos}} \) to \( \mathbf{s}_k^{\text{pos}} \). The solid arrow
shown the predicted rotation is rotated by $\theta$ from \( \mathbf{a}_k^{\text{rot}} \).}
   \label{fig:3}
\end{figure}

\noindent\textbf{Situation Grounding Module.}
Directly predicting the situation's absolute position \( \mathbf{s}^{pos} \) and rotation \( \mathbf{s}^{pos} \) in 3D space is challenging due to the complexity of estimating precise spatial coordinates and orientations. To simplify this task, we propose using anchor points derived from objects within the scene. Specifically, we treat each object as an anchor, utilizing its center coordinates \( \mathbf{a}_k^{\text{pos}} \) and rotation \( \mathbf{a}_k^{\text{rot}} \) as reference points. Consequently, we only need to predict the offset \( \Delta \mathbf{p}_k \in \mathbb{R}^3 \) from the anchor position to the situation position and the angular difference \( \theta_k \) between the anchor rotation and the situation rotation, as illustrated in Fig.~\ref{fig:3}. The blue circles and arrows shows the anchor \( \mathbf{a}_k^{\text{pos}} \) and \( \mathbf{a}_k^{\text{rot}} \). The greens are ground truth. The dotted line shows the offset from \( \mathbf{a}_k^{\text{pos}} \) to \( \mathbf{s}_k^{\text{pos}} \). The solid arrow
shows the predicted rotation is rotated by $\theta$ from \( \mathbf{a}_k^{\text{rot}} \). We set each \( \mathbf{a}_k^{\text{rot}} \) point to the center of the room. Beacuse estimating the front face of each object is a difficult problem.

For rotation prediction, we convert the regression problem into a classification problem. Considering rotations around the vertical axis (\ie, yaw rotation) as an example (see Fig.~\ref{fig:3}), we discretize the rotation angle into \( B \) bins ranging from \( -\pi \) to \( \pi \). The target angular difference \( \theta_k \) can then be represented by the bin index \( \hat{b}_k \) for anchor \( k \):

\begin{equation}
\hat{\theta}_k = -\pi + \frac{2\pi}{B} \left( \hat{b}_k + \frac{1}{2} \right). 
\end{equation}

We employ another MLP \( f_{\text{grd}} \) to predict the confidence score \( c_k \in [0, 1] \), the position offset \( \Delta \mathbf{p}_k \), and the rotation bin \( \hat{b}_k \). Here, \( \mathbf{h}_{\text{GRD}} \) is the hidden state of the grounding token from the LLM's output, and \( \mathbf{h}_k \) is the hidden state corresponding to the visual token \( \tilde{\mathbf{v}}_k \) of anchor \( k \):
\begin{equation} 
(c_k,\ \Delta \mathbf{p}_k,\ \hat{b}_k) = f_{\text{grd}}\left( [ \mathbf{h}_{\text{GRD}};\ \mathbf{h}_k ] \right),
\end{equation}
where \( [ \mathbf{h}_{\text{GRD}};\ \mathbf{h}_k ] \) denotes the concatenation of the two hidden states.
The predicted situation position for anchor \( k \) is then computed as:
\begin{equation} 
\hat{\mathbf{s}}^{\text{pos}}_k = \mathbf{a}_k^{\text{pos}} + \Delta \mathbf{p}_k. 
\end{equation}

During inference, we select the most confident anchor
$
k^\ast = \arg\max_k c_k,
$
and use its predictions for the situation position and rotation, \ie,
$
\hat{\mathbf{s}}^{\text{pos}} = \hat{\mathbf{s}}^{\text{pos}}_{k^\ast}$, $\hat{\theta} = \hat{\theta}_{k^\ast}$. Then, the predicted angular difference \( \hat{\theta} \) is mapped back to a quaternion representation for the situation rotation. Assuming the anchor's rotation \( \mathbf{a}_{k^\ast}^{\text{rot}} \) is known, the situation rotation \( \hat{\mathbf{s}}^{\text{rot}} \) is calculated by applying the rotation difference to the anchor's rotation:
\begin{equation} 
\delta \hat{\mathbf{s}}^{\text{rot}} = \left( 0,\ 0,\ \sin\left( \frac{\hat{\theta}}{2} \right),\ \cos\left( \frac{\hat{\theta}}{2} \right) \right),
\end{equation}

\begin{equation} 
\hat{\mathbf{s}}^{\text{rot}} = \mathbf{a}_{k^\ast}^{\text{rot}} \otimes \delta \hat{\mathbf{s}}^{\text{rot}},
\end{equation}
where \( \otimes \) denotes quaternion multiplication. By transforming the prediction task into estimating relative offsets and rotations with respect to anchor points, we reduce the complexity associated with directly predicting absolute positions and orientations in 3D space. 

\subsection{Training}

Our training process comprises three stages. In the first stage, we train the connector to align the point cloud features with the text embedding space. In the second stage, we train the situation grounding module using the situation grounding task, enhancing the LLM's situational awareness within 3D environments. In the final stage, we fine-tune the entire model using downstream instruction data to further improve its performance. Throughout all stages, we fine-tune the LLM using LoRA~\cite{hu2022lora}, and we utilize LLaMa 3.1~\cite{touvron2023llama} as the base LLM.

\noindent\textbf{Region Text Alignment.}
To achieve point cloud-text alignment, we map the point cloud to images, ensuring that we retain only the objects that are visible within the images. We also incorporate depth information to filter out object point clouds that are completely occluded in the corresponding view. This results in region-text pairs, linking specific regions of the point cloud to their corresponding textual descriptions. Compared to prior methods \cite{huang2023embodied, chen2024ll3da} that rely on scene captions, our approach simplifies the training process by reducing the number of objects, eliminating ambiguity in spatial relationships due to explicit viewpoint information, and increasing the number of training samples. A single scene can be divided into multiple regions, providing diverse contexts for training.

\noindent\textbf{Situation Grounding.}
After establishing the point cloud-text alignment, we proceed with the situation grounding. During training, we supervise only the anchors whose positions are within a distance threshold \( D \) of the ground truth situation position \( \mathbf{s}^{\text{pos}} \). Formally, the set of supervised anchors is given by:
\[
\mathcal{K} = \left\{ k\ \big|\ \left\| \mathbf{a}_k^{\text{pos}} - \mathbf{s}^{\text{pos}} \right\|_2 \leq D \right\}.
\]
For these anchors, the position loss is defined as:
\begin{equation}
\mathcal{L}_{\text{pos}} = \sum_{k \in \mathcal{K}} \left\| \hat{\mathbf{s}}^{\text{pos}}_k - \mathbf{s}^{\text{pos}} \right\|_2^2,
\label{eq:L_pos}
\end{equation}
where \( \hat{\mathbf{s}}^{\text{pos}}_k \) is the predicted situation position for anchor \( k \).
The rotation loss is given by:
\begin{equation}
\mathcal{L}_{\text{rot}} = - \sum_{k \in \mathcal{K}} \sum_{b=1}^B y_b \log p_{k,b}^{\text{rot}},
\label{eq:L_rot}
\end{equation}
where \( p_{k,b}^{\text{rot}} \) is the predicted probability for rotation bin \( b \) for anchor \( k \), and \( y_b \) is the one-hot encoding of the ground truth rotation bin corresponding to the angular difference between \( \mathbf{a}_k^{\text{rot}} \) and \( \mathbf{s}^{\text{rot}} \).
The confidence loss encourages the confidence scores \( c_k \) to reflect the anchors' proximity to the ground truth position:
\begin{equation}
\mathcal{L}_{\text{conf}} = \sum_{k} \left| c_k - \exp\left( -\alpha \left\| \mathbf{a}_k^{\text{pos}} - \mathbf{s}^{\text{pos}} \right\|_2 \right) \right|,
\label{eq:L_conf}
\end{equation}
with \( \alpha \) being a scaling factor controlling the decay rate. During inference, we select the most confident anchor \( k^\ast = \arg\max_k c_k \) and use its predictions for the situation position and rotation.

\noindent\textbf{Instruction Tuning.}
After training with situation awareness, we finetune the LLM on downstream 3D reasoning tasks.
The task answer loss is defined as the standard cross-entropy loss for language modeling:
\begin{equation}
\mathcal{L}_{\text{ans}} = - \sum_{i=1}^T \log P(a_i\ |\ a_{<i},\ \text{Input}),
\label{eq:L_ans}
\end{equation}
where \( T \) is the length of the task answer \( a \), and \( P(a_i\ |\ a_{<i},\ \text{Input}) \) is the probability of generating token \( a_i \).

%% file: sec/4_experiment.tex
\section{Experiments}

\begin{table*}[h]
\centering
\small
\label{tab:comparison}
\resizebox{0.9\textwidth}{!}{%
\begin{tabular}{lccccccccccc}
\toprule
\multirow{2}{*}{\textbf{Model}} & \multicolumn{5}{c}{\textbf{Scan2Cap (val)}} & \multicolumn{5}{c}{\textbf{ScanQA (val)}} & \textbf{SQA3D (test)} \\
\cmidrule(lr){2-6} \cmidrule(lr){7-11} \cmidrule(lr){12-12}
\textbf{} & \textbf{C} & \textbf{B-4} & \textbf{M} & \textbf{R} & \textbf{Sim} & \textbf{C} & \textbf{B-4} & \textbf{M} & \textbf{R} & \textbf{EM@1} & \textbf{EM@1} \\
\midrule
\textit{Task-specific models} & & & & & & & & & & \\
Scan2Cap \cite{chen2021scan2cap} & 35.2 & 22.4 & 21.4 & 43.5 & - & - & - & - & - & - & 41.0 \\
3DJCG \cite{cai_3djcg_2022} & 47.7 & 31.5 & 24.3 & 51.8 & - & - & - & - & - & - & - \\
Vote2Cap-DETR \cite{chen2023end} & 61.8 & 34.5 & 26.2 & 54.4 & - & - & - & - & - & - & - \\
ScanRefer+MCAN & - & - & - & - & - & 55.4 & 7.9 & 11.5 & 30.0 & 18.6 & - \\
ClipBERT & - & - & - & - & - & - & - & - & - & - & 43.3 \\
ScanQA \cite{azuma2022scanqa} & - & - & - & - & - & 64.9 & 10.1 & 13.1 & 33.3 & 21.1 & 47.2 \\
SIG3D\cite{man2024situational} & - & - & - & - & - & 68.8 & 12.4 & 13.4 & 35.9 & - & 52.6 \\
\midrule
\textit{Generalist models} & & & & & & & & & & \\
3D-VisTA \cite{zhu20233d} & 66.9 & 34.0 & 27.1 & 54.3 & 53.8 & 69.6 & 10.4 & 13.9 & 35.7 & 22.4 & 48.5 \\
3D-LLM (FlanT5) & 69.4 & 12.0 & 14.5 & 35.7 & - & - & - & - & - & - & - \\
LL3DA \cite{chen2024ll3da} & 65.2 & 36.8 & 26.0 & 55.1 & - & 76.8 & 13.5 & 15.9 & 37.3 & - & - \\
LEO \cite{huang2023embodied} & 72.4 & 38.2 & 27.9 & 58.1 & 55.3 & \textcolor{gray}{101.4} & \textcolor{gray}{13.2} & \textcolor{gray}{20.4} & \textcolor{gray}{49.2} & \textcolor{gray}{24.5 (47.6)} & 50.0 (52.4) \\
\textbf{Ours} & \textbf{75.2} & \textbf{38.9} & \textbf{29.0} & \textbf{58.7} & \textbf{56.3} & \textbf{89.8} & \textbf{14.6} & \textbf{17.5} & \textbf{42.9}  & \textbf{22.9 (40.2)} & \textbf{54.0 (56.0)} \\
\bottomrule
\end{tabular}%
}
\caption{Quantitative comparison with state-of-the-art models on 3D VL understanding tasks. “C” stands for “CIDEr”, “B-4” for “BLEU-4”, “M” for “METEOR”, “R” for “ROUGE”, “Sim” for sentence similarity, and “EM@1” for top-1 exact match. The n-gram metrics for Scan2Cap are governed by IoU@0.5. Entries in \textcolor{gray}{gray} indicate using ground truth question-relative object annotations.}
\end{table*}

\begin{table}[t]
\centering
\resizebox{\linewidth}{!}{%
\begin{tabular}{lcccc}
\toprule
\multirow{2}{*}{\textbf{Model}} & \multicolumn{2}{c}{\textbf{Localization}} & \multicolumn{2}{c}{\textbf{Orientation}} \\
\cmidrule(lr){2-3} \cmidrule(lr){4-5}
 & \textbf{Acc@0.5m} & \textbf{Acc@1.0m} & \textbf{Acc@15$^\circ$} & \textbf{Acc@30$^\circ$} \\
\midrule
Random & 7.2 & 25.8 & 8.4 & 16.9 \\
SQA3D \cite{ma2022sqa3d} & 9.5 & 29.6 & 8.7 & 16.5 \\
SQA3D \textit{(separate)} & 10.3 & 31.4 & 17.1 & 22.8 \\
3D-VisTA \cite{zhu20233d} & 11.7 & 34.5 & 16.9 & 24.2 \\
SIG3D$^\ast$ \cite{man2024situational} & 16.8 & 35.2 & 23.4 & 26.3 \\
\textbf{Ours} & \textbf{17.4} & \textbf{36.9} & \textbf{24.1} & \textbf{28.5} \\
\bottomrule
\end{tabular}%
}
\caption{Performance comparison of different models on localization and orientation metrics. $^\ast$ indicates our reproduction using their open-sourced code.}
\label{tab:sg}
\vspace{-2mm}
\end{table}

\subsection{3D Scene Understanding}
\noindent\textbf{Overview.}
We evaluate our method on three well-established 3D scene understanding tasks. Specifically, Scan2Cap \cite{chen2021scan2cap} requires the model to generate captions of each object in the scene regarding their category, attributes, and neighbor content. ScanQA \cite{azuma2022scanqa} requires the model to answer questions related to objects in 3D. SQA3D~\cite{ma2022sqa3d} requires the model to answer questions under particular situations described by the text. We investigate how well our method can perform 3D VL understanding and reasoning tasks, especially when compared against task-specific models~\cite{man2024situational, cai_3djcg_2022} and existing generalist models \cite{hong20233d, huang2023embodied}. We evaluate our model using conventional text generation metrics, including CIDEr~\cite{vedantam2015cider}, BLEU~\cite{papineni2002bleu}, METEOR~\cite{banerjee2005meteor}, and ROUGE-L~\cite{lin2004rouge}, and open-ended generation metric Sentence-Sim~\cite{reimers2019sentence} and refined exact-match accuracy~\cite{huang2023embodied}. Following 3D-VisTA~\cite{zhu20233d}, we utilize object proposals from Mask3D~\cite{schult2023mask3d} instead of ground-truth object segments for evaluation.


\noindent\textbf{Results and Analysis.} Existing methods for 3D scene understanding can be categorized into two streams: specialist models and generalist models. Specialist models are designed specifically for individual tasks only. Generalist models \ie LEO~\cite{huang2023embodied} allow for joint training and inference across different datasets without changing the network structure.
Compared with LEO, our method surpasses 2.8 CIDEr scores on Scan2Cap and 4\% on EM@1 on SQA3D. This improvement underscores the effectiveness of integrating 3D situational awareness into LLMs for enhancing 3D scene understanding and reasoning capabilities.

\subsection{Situation Grounding}
\noindent\textbf{Overview.}
In this part, we evaluate our method's ability to predict the position and orientation of agents based on textual descriptions using the SQA3D dataset. This dataset provides 26,000 situational descriptions, making it a comprehensive benchmark for 3D scene understanding from an egocentric perspective.
To assess our model's performance on situation grounding, we use the four metrics:
Acc@0.5m and Acc@1.0m, which are the percentages of position predictions within 0.5 meters and 1.0 meters of the ground truth on the x-y plane; Acc@15° and Acc@30°, which are the percentages of rotation predictions within 15 degrees and 30 degrees of the ground truth around the z-axis (yaw rotation).
The experiments in Table~\ref{tab:sg} demonstrate that our model effectively grounds textual descriptions into accurate spatial positions and orientations within 3D scenes. This highlights its potential for real-world applications where precise localization and orientation based on language inputs are crucial.

\subsection{Situation Captioning}
Describing the 3D environment from a first-person perspective is a critical task in embodied applications, such as navigation. In this experiment, we evaluate 3D LLMs on the task of generating region-level captions conditioned on a given situation. We start by using the agent's positional and rotational data, along with the camera position, to filter the point cloud to only the region visible to the agent. This filtered point cloud is then fed into the model to generate captions. The evaluation is conducted on 7,074 samples from the ScanNet validation set. We consider two types of captions: simple captions, which focus only on the main objects, and detailed captions, which capture the full environment in greater detail. The results, presented in Table~\ref{tab:view2cap}, show that LEO trained solely on object and scene-level captioning data perform poorly on situation captioning tasks. In contrast, our View2Cap data significantly improves performance. Additionally, training with both simple and detailed captioning data yields better results compared to using only one type of captioning data, suggesting that a diverse range of captioning styles can enhance model training and performance.

\begin{table}[t]
\centering
\resizebox{\linewidth}{!}{%
\begin{tabular}{lcccccc}
\toprule
\textbf{Type} & \textbf{Data} & \textbf{C} & \textbf{B-4} & \textbf{M} & \textbf{R} & \textbf{Sim} \\ \midrule
\multirow{4}{*}{Simple}  & SceneCap & 5.7 & 2.5 & 12.9 & 18.5 & 49.7 \\
 & SceneVerse & 5.7 & 2.5 & 12.9 & 18.5 & 49.7 \\
 & View2Cap (S) & 31.3 & 14.1 & 17.8 & 38.1 & 64.0 \\ 
 & View2Cap & \textbf{36.0} & \textbf{15.0} & \textbf{18.5} & \textbf{38.5} & \textbf{65.4} \\ 
\midrule
\multirow{4}{*}{Detail} & SceneCap & 1.5 & 0.5 & 8.6 & 15.8 & 50.9 \\
 & SceneVerse & 4.3 & 1.5 & 8.6 & 15.8 & 50.9 \\
 & View2Cap (D) & 11.2 & 11.8 & 19.1 & 29.8 & 69.0 \\ 
 & View2Cap & \textbf{12.5} & \textbf{12.2} & \textbf{19.9} & \textbf{29.9} & \textbf{70.0} \\ 
\bottomrule
\end{tabular}
}
\caption{Result on situation captioning training with different data.}
\label{tab:view2cap}
\vspace{-2mm}
\end{table}

\begin{figure*}[t]
  \centering
   \includegraphics[width=1.0\linewidth]{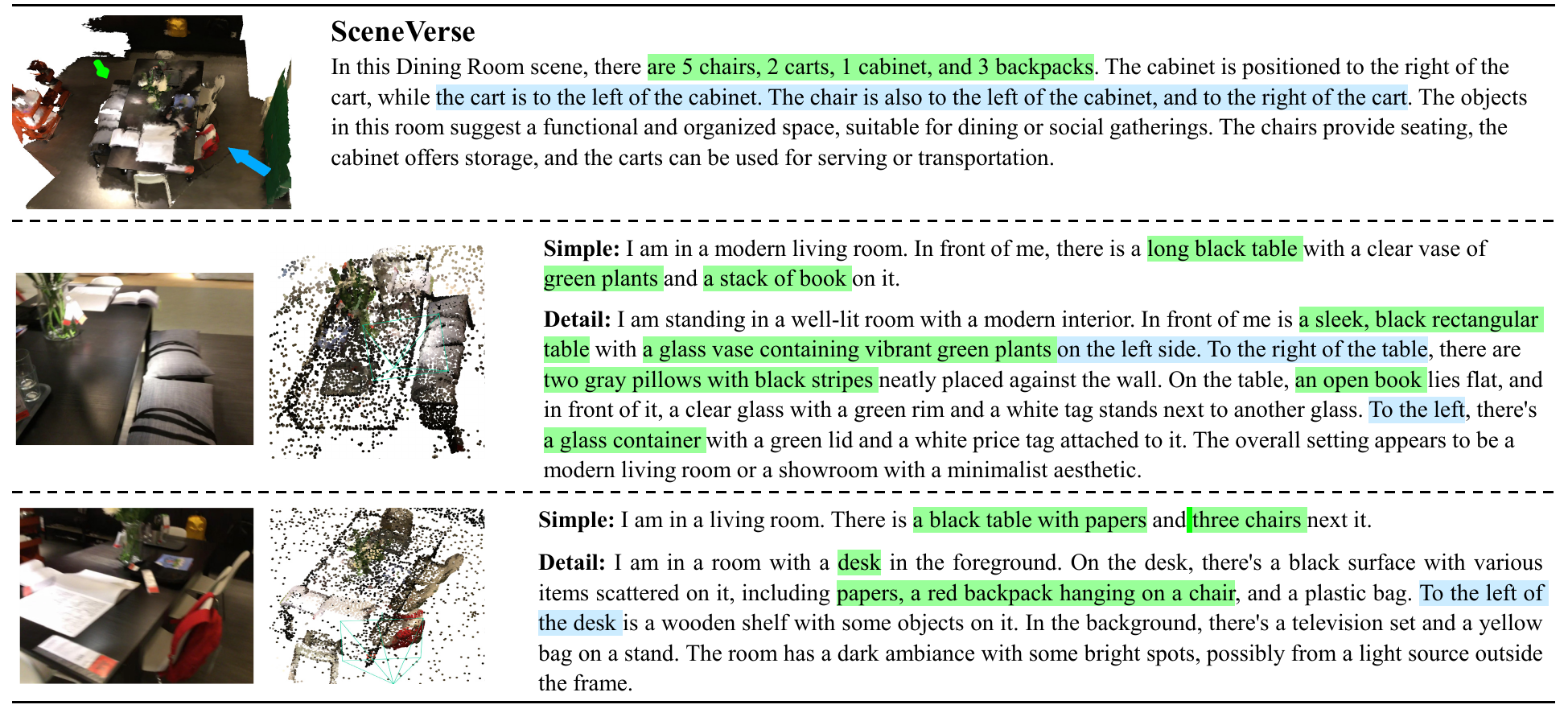}
   \vspace{-1.5em}
   \caption{Examples of our View2Cap situation captions against SceneVerse. We mark facts in \colorbox[RGB]{153, 255, 153}{green} and spatial relations in \colorbox[RGB]{204, 236, 255}{blue}.}
   \label{fig:4}
   \vspace{-10pt}
   
\end{figure*}

\subsection{Situation Question Answering}
We evaluate our model's performance on the situational VQA task using our generated ViewQA dataset. For the single-view QA, we input the region point cloud into the model, similar to the procedure in the situational captioning task. For the multi-view QA, we provide the model with a scene-level point cloud along with a situational description that includes the starting point and actions as conditions. The model is then tasked with answering questions about the differences or new views resulting from this action. The validation set comprises 8,287 questions derived from the ScanNet dataset. As shown in Table~\ref{tab:abl_data}, incorporating the Situation Grounding (SG) Module for situational prediction and utilizing the View2Cap data for region-text alignment significantly improves performance on our ViewQA task.

\subsection{Qualitative Analysis}
Fig.~\ref{fig:4} presents the qualitative example of our View2Cap dataset complete against scene graph based generation method SceneVerse~\cite{jia2024sceneverse}. The first column shows the SceneVerse caption, it just summarizes the objects in the scene and describes the relation of objects in predefined rules. While our method gives more detailed object descriptions and accurate spatial relationships. For instance, the caption from SceneVerse ignores the glass vase and opened books on the table as our View2Cap describes.

\begin{table}[t]
\centering
\resizebox{\linewidth}{!}{%
\begin{tabular}{lccccc}
\toprule
 & \multicolumn{2}{c}{\textbf{Localization}} & \multicolumn{2}{c}{\textbf{Orientation}} \\
\cmidrule(lr){2-3} \cmidrule(lr){4-5}
 & \textbf{Acc@0.5m} & \textbf{Acc@1.0m} & \textbf{Acc@15$^\circ$} & \textbf{Acc@30$^\circ$} \\
\midrule
LEO w. SG & 8.3 & 30.4 & 10.9 & 19.5\\
+ Anchor & 13.7 & 32.2 & 16.9 & 21.8 \\
+ Discrect bins & 13.6 & 32.3 & 21.6 & 25.0 \\
+ View2Cap & \textbf{17.4} & \textbf{36.9} & \textbf{24.1} & \textbf{28.5} \\
\bottomrule
\end{tabular}%
}
\vspace{-0.5em}
\caption{Ablations of our designs on situation grounding.}
\label{tab:abl_sg}
\vspace{-1em}
\end{table}

\subsection{Ablation Study}

\noindent\textbf{Design of Situation Grounding Module.}
We test different designs of situation grounding modules as shown in Table~\ref{tab:abl_sg}. Compared with the model without anchors, the localization Acc@1.0m improves 5.4\%, indicating that the anchor mechanism can help the model narrow down the situation in a smaller range. Additionally, using the discrete bins to predict the rotation provides more accurate angles. Pretraining using View2Cap on the captioning task can improve both the position and rotation performance.

\noindent\textbf{Effectiveness of Situation Data.}
To evaluate the effectiveness of situation data, we test the performance of the model on SQA3D, as it requires both the situation grounding and QA ability. We also test the influence of situation data on 3D visual grounding, as it also needs to infer the situation to distinguish distractors. The result in Table~\ref{tab:abl_data} shows that pretraining on our proposed View2Cap and situation grounding can effectively improve the performance on those tasks that need situation awareness.

\begin{table}[t]
\centering
\resizebox{\linewidth}{!}{%
\begin{tabular}{lccccccc}
\toprule
 & \multicolumn{2}{c}{\textbf{ViewQA}} & \multicolumn{2}{c}{\textbf{SQA3D}} & \multicolumn{2}{c}{\textbf{ScanRefer}} \\
\cmidrule(lr){2-3} \cmidrule(lr){4-5} \cmidrule(lr){6-7}
 & \textbf{EM} & \textbf{EM-R} & \textbf{EM} & \textbf{EM-R} & \textbf{Acc@0.25} & \textbf{Acc@0.5} \\
\midrule
LEO  & 39.3 & 44.1 & 62.8 & 52.4 & 36.1 & 30.8 \\
+ SG module & 40.2 & 45.3 & 50.8 & 53.2 & 38.3 & 32.9 \\
+ View2Cap & \textbf{42.0} & \textbf{46.6} & \textbf{54.0} & \textbf{56.0} & \textbf{42.8} & \textbf{38.4} \\
\bottomrule
\end{tabular}%
}
\vspace{-0.5em}
\caption{Ablations of our designs on situation-aware reasoning.}
\label{tab:abl_data}
\vspace{-1em}
\end{table}

%% file: sec/5_conclusion.tex
\section{Conclusion}

In this paper, we presented a novel approach to enhance 3D LLMs with situational awareness. Recognizing the limitations of existing methods that overlook the egocentric perspective inherent in 3D environments, we proposed the automatic generation of a situation-aware dataset called View2Cap. By leveraging the scanning trajectories from RGB-D video data and utilizing powerful VLMs, we produced high-quality captions and QA pairs that capture the dynamic viewpoints of an observer moving through a 3D scene. Furthermore, we introduced a situation grounding module, enabling LLMs to ground textual descriptions to situations in 3D space explicitly. We hope our work will advance the first-person 3D understanding of embodied tasks.
\clearpage

%% file: sec/6_ac.tex
\section{Acknowledgements}

This work was supported by NSFC with Grant No. 62293482, by the Basic Research Project No. HZQB-KCZYZ-2021067 of Hetao Shenzhen HK S\&T Cooperation Zone, by Shenzhen General Program No. JCYJ20220530143600001, by Shenzhen-Hong Kong Joint Funding No. SGDX20211123112401002, by the Shenzhen Outstanding Talents Training Fund 202002, by Guangdong Research Project No. 2017ZT07X152 and No. 2019CX01X104, by the Guangdong Provincial Key Laboratory of Future Networks of Intelligence (Grant No. 2022B1212010001), by the Guangdong Provincial Key Laboratory of Big Data Computing, CHUK-Shenzhen, by the NSFC 61931024\&12326610, by the Key Area R\&D Program of Guangdong Province with grant No. 2018B030338001, by the Shenzhen Key Laboratory of Big Data and Artificial Intelligence (Grant No. ZDSYS201707251409055), by Shaanxi Mathematical Basic Science Research Project (No.23JSY047), and by Tencent \& Huawei Open Fund, by China Association for Science and Technology Youth Care Program.

%% file: sec/X_suppl.tex
\clearpage
\setcounter{page}{1}
\maketitlesupplementary

\begin{figure*}[t]
  \centering
   \includegraphics[width=0.9\linewidth]{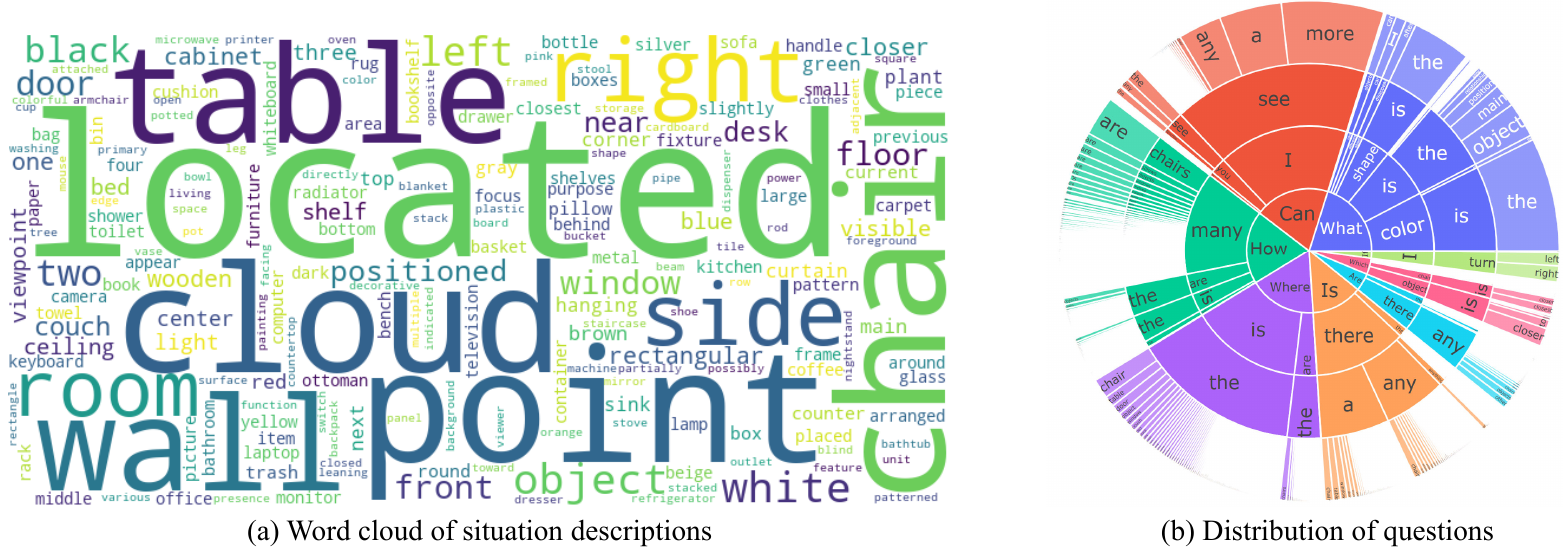}
   \caption{Dataset statistics.}
   \label{fig:stat_1}
   
\end{figure*}

\begin{figure*}[t]
  \centering
   \includegraphics[width=0.9\linewidth]{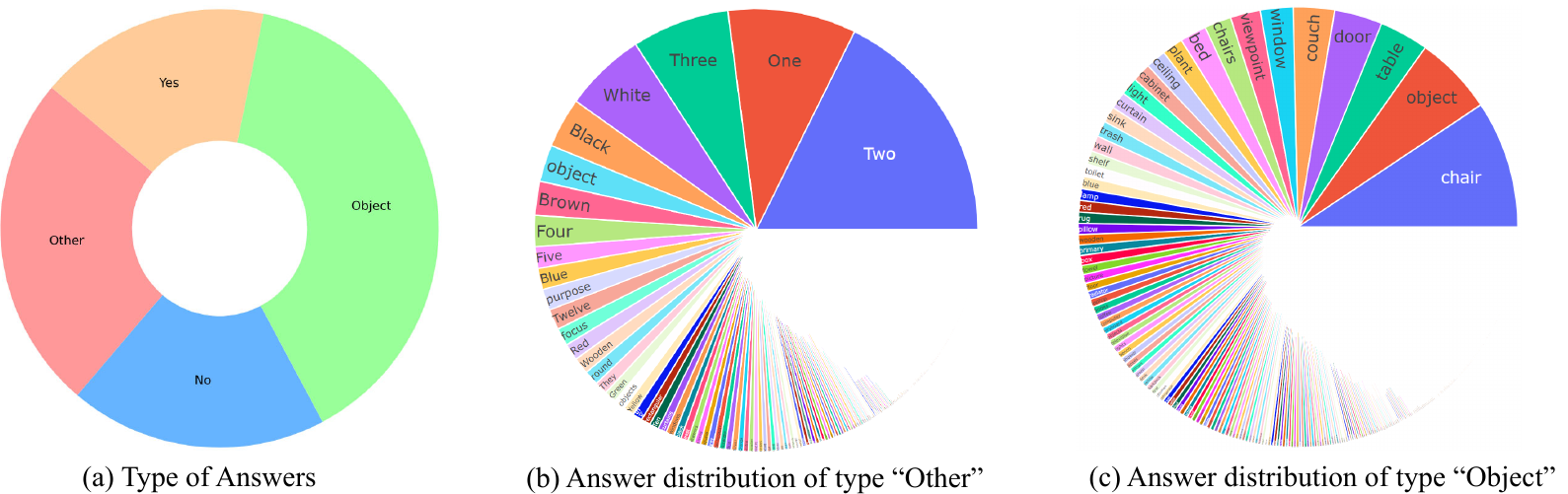}
   \caption{Answer distributions.}
   \label{fig:stat_2}
   
\end{figure*}

\begin{figure*}[t]
  \centering
   \includegraphics[width=0.9\linewidth]{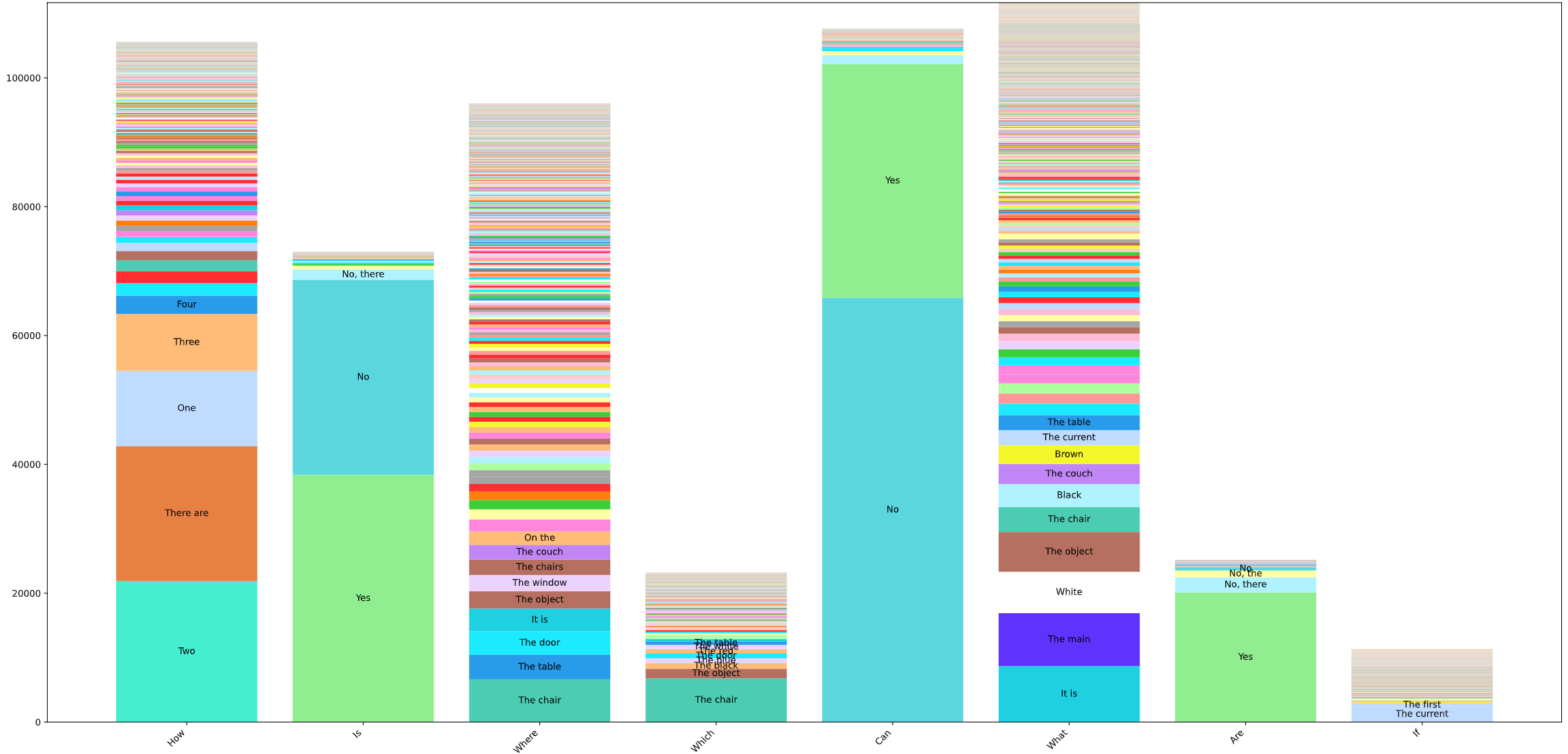}
   \caption{Answer distribution organized by question prefixes.}
   \label{fig:answer bar}
   
\end{figure*}

\section{Dataset Statistics}
\label{sec:supp_stat}
We present the detailed statistics of our dataset. The View2Cap dataset comprises 231,184 situation descriptions and 553,779 question answers, paired with corresponding viewpoint positions and rotations. Fig.~\ref{fig:stat_1}(a) illustrates the word cloud of the situation descriptions, where the size of each word indicates its frequency in the dataset. Prominent words such as ``located," ``cloud," ``point," ``table," ``wall," and ``light" appear in larger fonts, while less frequent words like ``metal," ``lamp," ``shelf," and ``keyboard" are smaller. Many terms represent spatial descriptions or object placements, including ``left," ``right," ``window," and ``front." Fig.~\ref{fig:stat_1}(b) displays the distribution of questions based on their prefixes, demonstrating the diversity of our questions, which encompass various inquiries about object attributes, quantities, and spatial relationships.

Fig.~\ref{fig:stat_2} presents the distribution of answers categorized by their types and contents. The answers are divided into four types: ``Yes," ``No," ``Object," and ``Other." Fig.~\ref{fig:stat_2}(b) visualizes the distribution of answers in the ``Other" category, showcasing a wide range of specific values such as colors and numbers. Fig.~\ref{fig:stat_2}(c) displays the distribution of answers within the ``Object" category, emphasizing the variety of object types mentioned, with notable contributions from common items such as ``chair," ``table," and ``door." These visualizations provide a comprehensive overview of the types of answers in the dataset and their respective distributions, reflecting the semantic diversity presented in the data.

Fig.~\ref{fig:answer bar} visualizes the diversity and frequency of answers corresponding to various question categories such as "How," "Is," "Where," "What," and "Can." Each stacked bar represents the range of unique responses, with more frequent answers prominently displayed at the base of the stack. Notable patterns include the dominance of simple affirmative ("Yes") or negative ("No") answers in certain categories (e.g., "Is" and "Can") and the high variability in object-related responses under "What" and "Where." This visualization underscores the semantic richness and heterogeneity of the dataset while highlighting common response tendencies for specific question types.

\section{Dataset Refinement Process}
\begin{figure*}[t]
  \centering
   \includegraphics[width=1.0\linewidth]{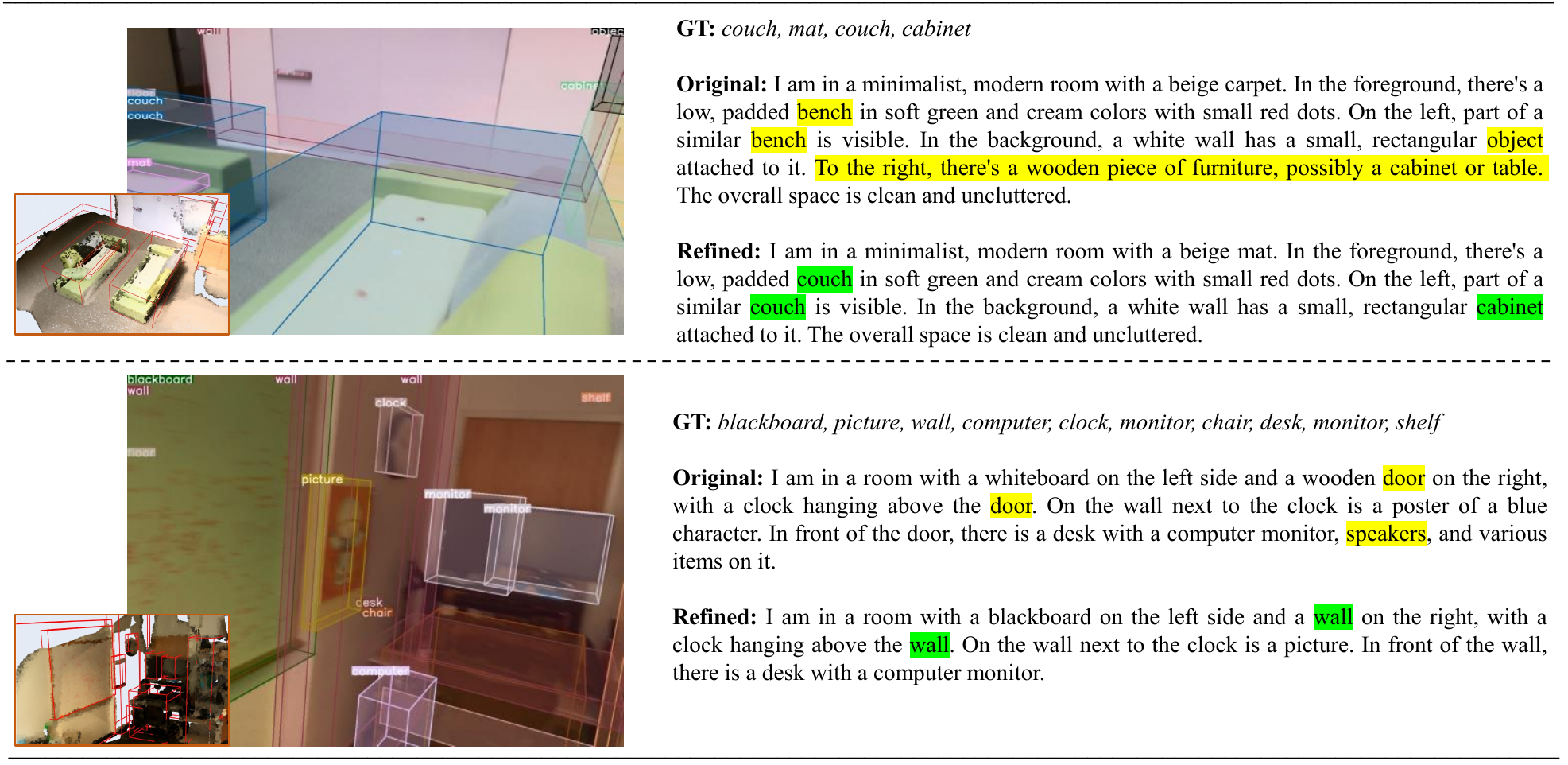}
   \caption{Examples of our View2Cap captions.}
   \label{fig:refine_1}
   
\end{figure*}

We show the details of using human-annotated object labels to refine the View2Cap dataset. After generating image caption using Llava-onevision~\cite{li2024llava}, we feed the original caption and ground truth 3D object labels from EmbodiedScan~\cite{wang2024embodiedscan} to GPT4o and ask it to score the caption and refine it. In Fig.~\ref{fig:refine_1}, we show the image and region point cloud on the left. On the right, we show the ground truth labels, the original caption, and the refined caption. It can be seen that our original caption can capture more fine-grained objects and detailed information than 3D labels, while refinement, can further improve the correctness. 

\begin{figure*}[t]
  \centering
   \includegraphics[width=1.0\linewidth]{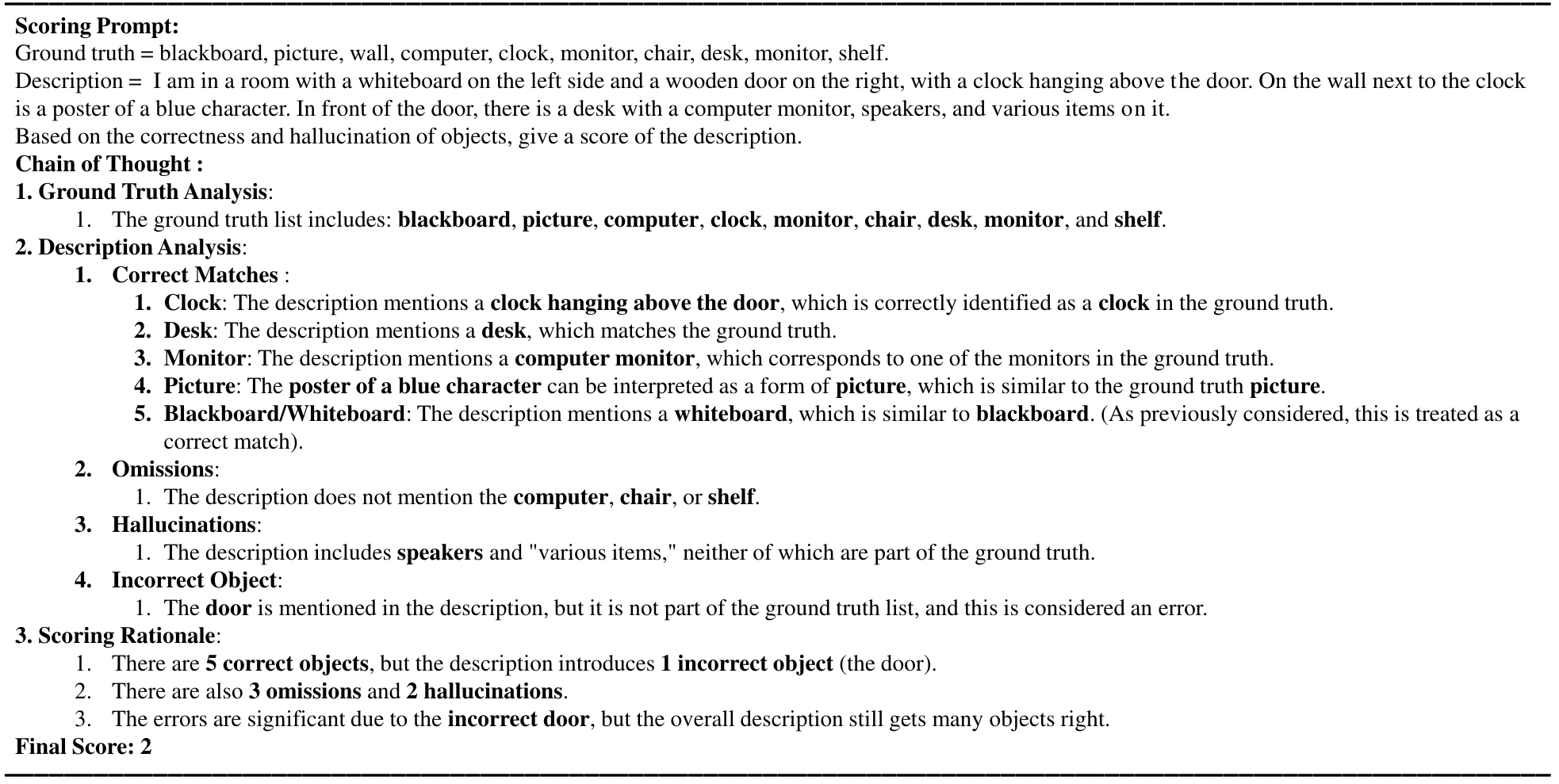}
   \caption{Example of our View2Cap verification process.}
   \label{fig:refine_2}
   
\end{figure*}

\begin{figure*}[t]
  \centering
   \includegraphics[width=1.0\linewidth]{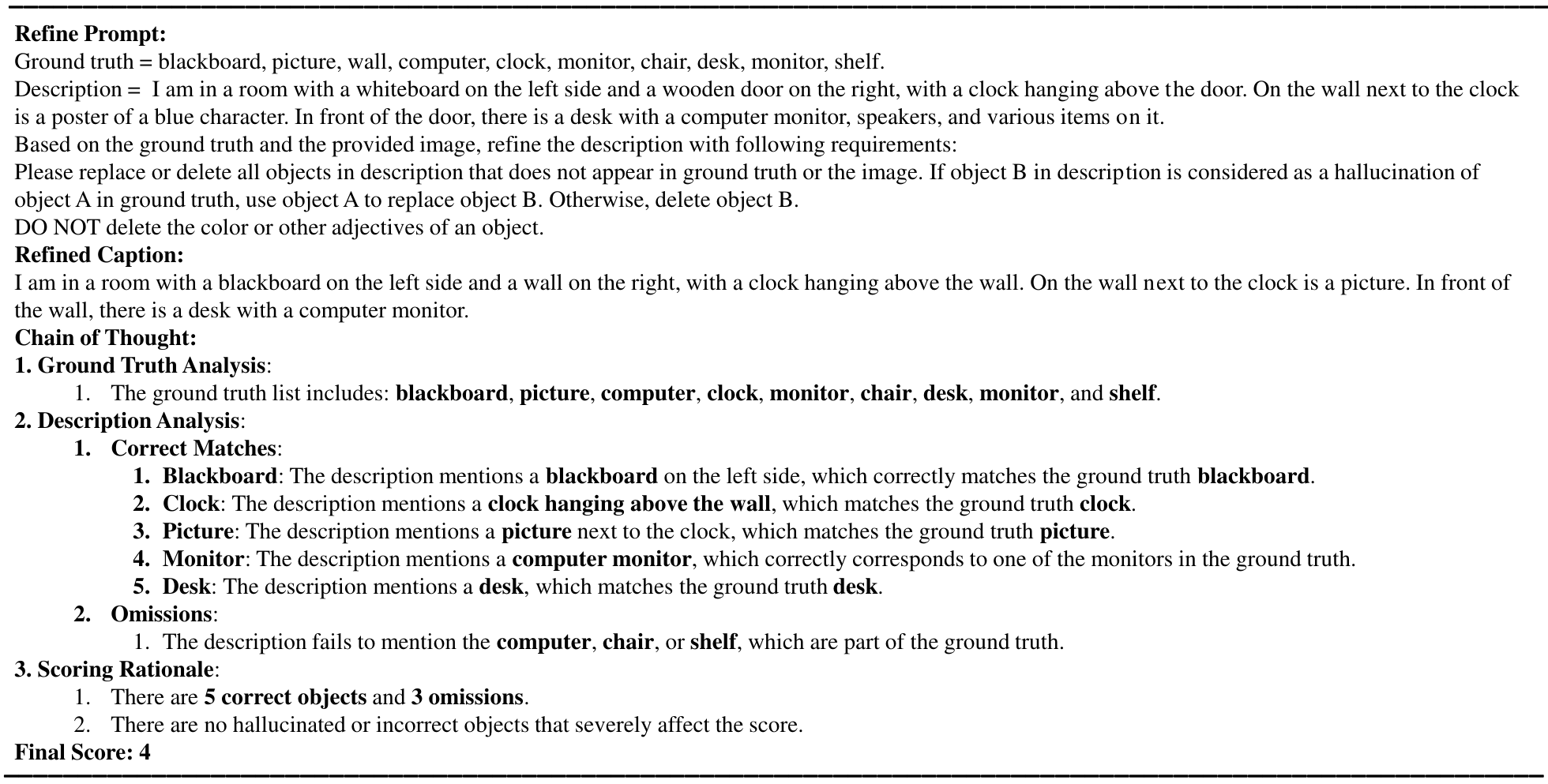}
   \caption{Example of our View2Cap refinement process.}
   \label{fig:refine_3}
   
\end{figure*}

In Fig.~\ref{fig:refine_2} and Fig.~\ref{fig:refine_3}, we demonstrate a structured approach for evaluating and refining a generated caption against a given ground truth. The process includes a detailed scoring prompt, chain of thought analysis for identifying correctness and hallucinations, and a scoring rationale leading to a final score. The refinement prompt outlines specific instructions to correct hallucinated or omitted objects, ensuring alignment with the ground truth. The refined caption achieves a higher score by eliminating errors and improving descriptive accuracy.

\section{Implementation Details}
We set the maximum input token length and output new token length of Llama 3.1 7B~\cite{touvron2023llama} to 256. For each 3D scene, we sample up to 60 instances. For each instance, the maximum point number is 1024. During the training, the point cloud encoder is frozen, and the LLM is fine-tuned with Low-Rank Adaptation (LoRA) super-parameters rank and alpha 16. We add LoRA parameters on all linear layers. The learning rate is $2e^{-4}$, and the batch size is 8 on each GPU. We use 4 NVIDIA A100 GPUs with 80G memory in all experiments. During inference, we use the beam search strategy to generate text. The beam size is set to 5.

